METHODOLOGY

# Active learning with point supervision for cost-effective panicle detection in cereal crops

Akshay L. Chandra[1], Sai Vikas Desai[1], Vineeth N. Balasubramanian[1*], Seishi Ninomiya[2] and Wei Guo[2*]

## Abstract

**Background:** Panicle density of cereal crops such as wheat and sorghum is one of the main components for plant breeders and agronomists in understanding the yield of their crops. To phenotype the panicle density effectively, researchers agree there is a significant need for computer vision-based object detection techniques. Especially in recent times, research in deep learning-based object detection shows promising results in various agricultural studies. However, training such systems usually requires a lot of bounding-box labeled data. Since crops vary by both environmental and genetic conditions, acquisition of huge amount of labeled image datasets for each crop is expensive and time-consuming. Thus, to catalyze the widespread usage of automatic object detection for crop phenotyping, a cost-effective method to develop such automated systems is essential.

**Results:** We propose a point supervision based active learning approach for panicle detection in cereal crops. In our approach, the model constantly interacts with a human annotator by iteratively querying the labels for only the most informative images, as opposed to all images in a dataset. Our query method is specifically designed for cereal crops which usually tend to have panicles with low variance in appearance. Our method reduces labeling costs by intelligently leveraging low-cost weak labels (object centers) for picking the most informative images for which strong labels (bounding boxes) are required. We show promising results on two publicly available cereal crop datasets—Sorghum and Wheat. On Sorghum, 6 variants of our proposed method outperform the best baseline method with more than 55% savings in labeling time. Similarly, on Wheat, 3 variants of our proposed methods outperform the best baseline method with more than 50% of savings in labeling time.

**Conclusion:** We proposed a cost effective method to train reliable panicle detectors for cereal crops. A low cost panicle detection method for cereal crops is highly beneficial to both breeders and agronomists. Plant breeders can obtain quick crop yield estimates to make important crop management decisions. Similarly, obtaining real time visual crop analysis is valuable for researchers to analyze the crop's response to various experimental conditions.

## Background

The widespread success of deep learning has spawned a multitude of applications in computer vision based plant phenotyping. State-of-the-art convolutional neural networks have been shown to perform well on a wide variety of phenotyping tasks. The applications of CNNs in plant phenotyping include image classification tasks such as plant species identification [1], stress identification [2], object detection and counting tasks such as panicle or spike detection [3–7], leaf counting [8], fruit detection [9]; as well as pixel-wise segmentation based tasks such as panicle segmentation [10, 11] and crop-weed segmentation [12]. We refer the reader to [13, 14] for a full treatment of deep learning in agriculture and plant phenotyping tasks.

*Correspondence: vineethnb@iith.ac.in; guowei@g.ecc.u-tokyo.ac.jp
[1] Department of Computer Science and Engineering, Indian Institute of Technology Hyderabad, Kandi, Sangareddy 502285, India
[2] International Field Phenomics Research Laboratory, The University of Tokyo, Graduate School of Agricultural and Life Sciences, Nishi-Tokyo, Tokyo 1880002, Japan

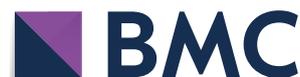

Despite many studies showing success in plant phenotyping tasks, the practical usage of deep learning in plant phenotyping poses a fundamental problem: requirement of large labeled datasets. Depending on the complexity of the phenotyping task and desired accuracy, large training sets may be needed to train deep learning models. However, there is a scarcity of publicly available agricultural image datasets. Since plant phenotyping tasks can be very specific to certain environmental and genetic conditions, finding labeled datasets with exact such conditions is often very difficult. This results in the researchers having to acquire and curate their own datasets which is a time-consuming and expensive task. Due to the emergence of services such as Amazon Mechanical Turk (AMT),[1] crowd-sourcing annotations has evolved to become a low cost solution to address the issue of efficiently creating large scale visual datasets. Crowd-sourcing [15–21] has been effectively used to generate datasets to train deep learning models for visual tasks. Vijayanarasimhan and Grauman [18] have attempted to combine the advantages of crowd sourcing and active learning to train models with minimal amount of supervision. However, to reap the benefits of crowd-sourcing, effective planning and sufficient quality control measures are required [22, 23], which constitutes additional overhead. In our work, we focus on reducing annotation time using active learning. Moreover, our work can be seamlessly extended to work with a crowd-sourcing platform to obtain bounding box annotations.

In this paper, we focus on panicle detection in cereal crop images. Efficient panicle detection models greatly assist cereal crop phenotyping since they provide quick panicle count estimates which can be used for yield estimation. High throughput yield estimation methods are highly beneficial for both agronomists and breeders. Crop breeders will potentially make effective selection in large scale breeding programs. Also, real time yield estimation techniques can be used for crop monitoring during controlled crop experiments in various genetic and environmental conditions. However, such panicle detection models require a lot of labeled data to train, which makes these methods less applicable for new crops for which datasets are not available. To address this problem, we employ point supervision based active learning to reduce the number of labeled samples to train efficient detection models.

Point supervision is a form of weak supervision which has been used for training deep neural networks for tasks such as object detection [24, 25], semantic segmentation [25, 26] and object tracking [27, 28]. It involves pointing to objects of interest on an image using mouse clicks. Point supervision is significantly inexpensive and less time-taking to obtain, when compared to the conventional full supervision methods such as bounding box drawing and pixel-wise image labeling. In the context of object detection, point supervision can provide valuable information about the location of objects. Such location information can typically be used to train models with novel loss functions or incorporate multiple instance learning techniques. In terms of obtaining point supervision, the closest literature to our work is Papadopoulos et al. [24]. They propose a method to integrate point supervision to a Multiple Instance Learning (MIL) approach to train per-class SVMs for each object class, which is generally used for Weakly Supervised Object Localization (WSOL) [29, 30]. Similar to [24], we obtain point supervision in the form of object center clicks. However, instead of directly training an object detector using point supervision alone, we use it in conjunction with an active learning approach. Using point supervision alone limits the performance of the model and results in weak-learners since accurate labels are not provided to the model. Hence, we employ point supervision to help choose the best samples for which full supervision can be queried. To the best of our knowledge, our work is the first to use point supervision to assist an active learning approach to pick the most informative samples for labeling. In our paper, we use the terms *weak labels* to denote labels obtained through point supervision and *strong labels* to denote labels obtained through full supervision.

Active learning [31], an iterative training approach that curiously selects the best samples to train, has been shown to reduce labeled data requirement when training deep classification networks [32–34]. Research in the area of active learning for object detection [18, 35, 36] has been limited. These efforts propose various metrics to compute on the unlabeled data that help pick the best subsets to be labeled. However, they show results on standard public datasets like PASCAL VOC [37] and MS COCO [38]. In this study, we focus on object detection for agricultural crop datasets which have a few important differences from standard object detection datasets such as PASCAL VOC or MS COCO: (1) objects, generally, are of a single class or just a few classes, (2) number of objects per image are often high (25–100+), (3) objects can be under heavy occlusion due to factors like surrounding leaves, weed, shadows etc; and (4) background can often look like the foreground (e.g. green color). Owing to these factors, labeling crop images is tricky and time-consuming. Taking into account these important differences, we propose an active learning framework based on point supervision to reduce annotation efforts for panicle detection in crop images. We measure annotation cost in terms of time taken. Weakly supervised annotations such as object-center clicks take significantly less time to obtain when compared to regular bounding box annotations. These clicks provide valuable localization cues to the object detection model in our framework. Formally, we define two forms of image annotation: (i) object center clicking (Type-1) and (ii) bounding box drawing (Type-2). To select the best subset of images to annotate, we incorporate point supervision into our active learning query function, which has never been done before to the best of our knowledge. In our setting, we train our model in a slightly varied version of standard pool-based active learning (see Figs. 1 and 2) where we obtain weak labels of the images samples from unlabeled pool and maintain a separate weak labeled pool. Our experiments show that using affordable-to-obtain weak labels can be used to create better query functions to find out the most informative samples, leading to a reduction in annotation costs. Our methodology can be seamlessly extended to any crop detection task other than panicle detection.

---
[1] Amazon Mechanical Turk is available at https://www.mturk.com/.

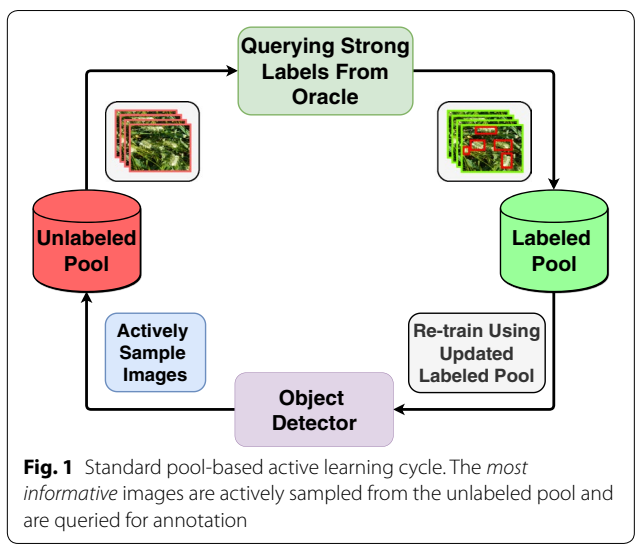

**Fig. 1** Standard pool-based active learning cycle. The *most informative* images are actively sampled from the unlabeled pool and are queried for annotation

## Methodology

### Annotation methods

Throughout our experiments, our image annotator a.k.a oracle provides annotations of objects of interest in images in two ways. These two methods differ in terms of the label quality and cost (in time units). Estimation of labeling costs of these methods is discussed in "Experimental setup" section. See Fig. 3 for visual illustration of these methods.

*Type-1 annotation* For each object in a given image, the oracle clicks approximately on the center of the imaginary bounding box that encloses the object. Since we obtain the center of each bounding box but not its dimensions, the label quality in Type-1 annotation is weak. The annotation cost in this case, is lower than that of a Type-2 annotation.

*Type-2 annotation* Given an image and its Type-1 annotations (weak labels), the oracle provides bounding boxes that tightly enclose the objects present. The labels in this case are strong since we get tight bounding boxes.

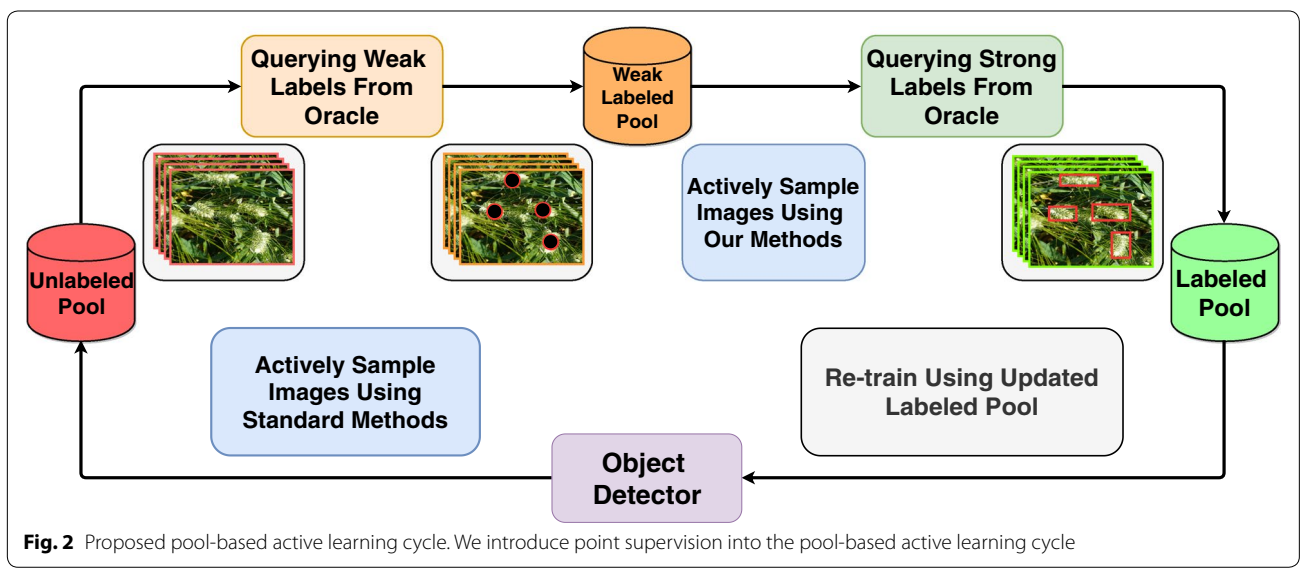

**Fig. 2** Proposed pool-based active learning cycle. We introduce point supervision into the pool-based active learning cycle

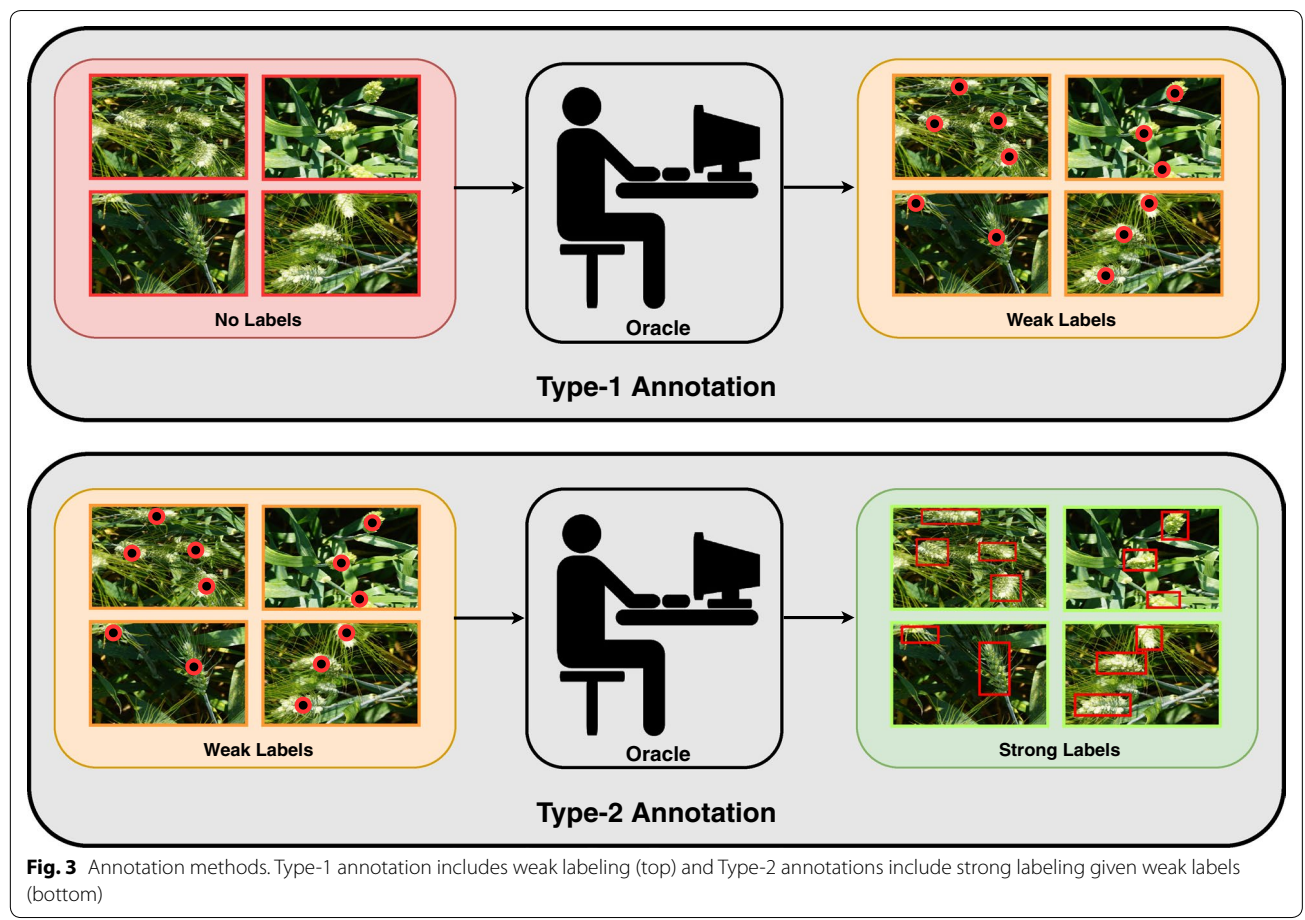

**Fig. 3** Annotation methods. Type-1 annotation includes weak labeling (top) and Type-2 annotations include strong labeling given weak labels (bottom)

Since the oracle is already given the weak labels, the annotation cost for Type-2 labels is lesser than the annotation cost for drawing bounding boxes from scratch. This is because the weak labels i.e., the object centers guide the oracle in locating the objects, thereby reducing the annotation cost.

**Overview of two stage object detection**

We use Faster R-CNN [39] as our chosen object detector in all our experiments. Faster R-CNN belongs to the family of two stage object detection networks. Two stage networks typically use a coarse to fine prediction approach. The first stage consists of generating possible object proposals on a given image. The second stage consists of (i) detecting the presence of an object within each proposal, (ii) classifying the object into one of the available categories and (iii) adjusting the proposals dimensions to tightly enclose the object. In Faster R-CNN, the first stage is implemented using a CNN based image level feature extractor (which is known as the backbone network) and a Region Proposal Network (RPN). First, the input image is passed through the base network to obtain image level features. Subsequently, the region proposal network (RPN) generates possible object proposals. The second stage consists of using RoI pooling [39] to obtain fixed dimension feature vectors for each object proposal. Each of these feature vectors is passed through a fully connected neural network with two heads: one for predicting the object class and the other for adjusting the bounding box coordinates. We use two stage object detection in our work because the output of RPN i.e., the first stage gives valuable information about model uncertainty which we use for efficient sample selection in active learning.

**Standard pool-based active learning framework**

The key assumption behind active learning is that a machine learning algorithm can achieve greater accuracy with fewer training labels if it is allowed to choose the data from which it learns. An active learner may pose queries, usually in the form of unlabeled data instances to be labeled by an oracle (for instance, a human annotator). Active learning is well-motivated in many modern machine learning problems, where unlabeled data may be abundant or easily obtained, but labels are difficult, time-consuming and expensive to obtain. Active learning involves a class of methods that are used to train machine learning models with limited labeled data by carefully picking the most valuable data points to be labeled. In case of deep neural networks, active learning is generally implemented under a setting known as pool-based active learning, see Fig. 1. This typically consists of the following five components: (1) model, (2) labeled pool of data, (3) unlabeled pool of data, (4) an active query function that samples data points from the unlabeled pool and (5) an oracle which provides labels when queried. The model is trained in cycles as follows: First, the model is trained on the available labeled pool. Using the model and unlabeled pool as input, the query function calculates an *informativeness measure* for each data point in the unlabeled pool and greedily samples the most informative data points. Labels for these points are obtained from the oracle, following which, these points are moved from the unlabeled pool to the labeled pool. Now, the model is retrained on the updated labeled pool and this process is repeated in iterations until the model converges to a desirable performance or until the annotation budget is exhausted.

Various techniques [31] have been proposed to calculate informativeness measures effectively. One popular technique is to estimate the model uncertainty on each data point. So the motivation is to pick the data points for labeling that confuse the model i.e. which have high model uncertainty and further not pick the data points on which the model is already confident about. In this paper, we propose a novel way of estimating the uncertainty of the model on images using their weak labels, in an object detection setting. Also, we modify the standard pool based active learning setting by obtaining weak labels instead of strong labels first and then make better *active* decisions on which data points to pick for strong labeling.

### Active learning with point supervision

The primary contribution of our paper is a novel method that incorporates point supervision to query uncertain images. To this end, we introduce a weak labeled pool into the standard pool-based active learning framework for training a deep object detector as shown in Fig. 2. Our method is designed for region proposal based two stage object detection networks such as Faster R-CNN and Mask R-CNN which usually have superior detection performance. Given an object detection model and weakly labeled pool of images, our query method takes the following steps: (1) Region proposal filtering using weak labels and (2) Estimate uncertainty using region proposals. The subsequent steps are similar to standard pool-based active learning. In other words, images with high uncertainty are picked by the proposed query function and strong labels are queried. Later, the labeled images are added to the labeled pool on which our object detection model is trained. This model is used in the next cycle of active learning (see Algorithm 1). Detailed description of the steps in our proposed query method are given in the following subsections.

---

**Algorithm 1:** Active Learning With Point Supervision

**Input:** Initial labeled pool $D^L$, unlabeled pool $D^U$, weak labeled pool $D^W$, model $\theta$, budget $B$, uncertainty function $U$, batch sizes $b_W$, $b_S$
**Output:** A model $\theta$ trained on the *most informative* samples

$\theta \leftarrow \texttt{Train}(D^L)$
**while** $B$ *and* $D^U$ *not exhausted* **do**
　　// Traditional active sampling
　　$Q \leftarrow \texttt{ActiveSampling}(\theta, D^U)$ $b_W$ images
　　$D^W \leftarrow D^W \cup \texttt{TypeOneAnnotation}(Q)$
　　$D^U \leftarrow D^U - Q$

　　// Our proposed active sampling which uses point supervision, see Algorithm 2.
　　$Q \leftarrow \texttt{ActiveSamplingPS}(\theta, D^W)$ $b_S$ images
　　$D^L \leftarrow D^L \cup \texttt{TypeTwoAnnotation}(Q)$
　　$D^W \leftarrow D^W - Q$

　　// Re-train model on newly updated labeled pool.
　　$\theta \leftarrow \texttt{Train}(D^L)$
　　$B \leftarrow B - b_W - b_S$

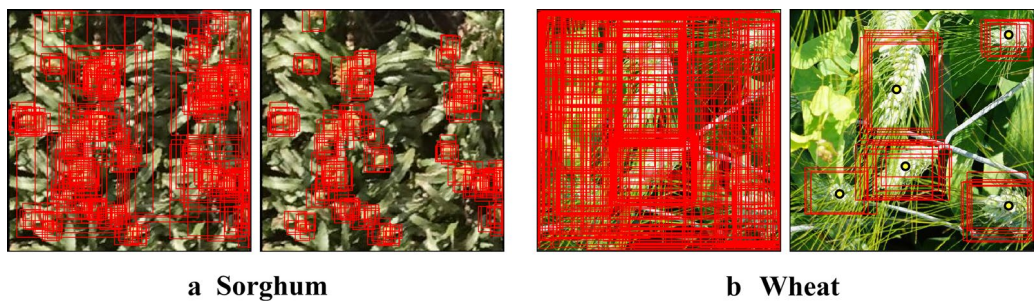

**Fig. 4** Region proposal filtering. Before filtering-after filtering illustration on **a** Sorghum and **b** Wheat. For ease of clarity, weak labels (yellow dots) are only shown on the **b** Wheat example

### *Region proposal filtering (RPF)*

We run the model on each image in the weakly labeled pool and obtain a set of region proposals from the Region Proposal Network (RPN). We now incorporate the point supervision signal i.e., the click annotations to filter out spurious region proposals from the set (see Fig. 4). For a proposal to be retained, (1) it must contain the click location, (2) it must have its center within $\varepsilon$ distance from the click location and (3) its area should not exceed a threshold $\alpha$. Here, $\varepsilon$ and $\alpha$ are hyperparameters which we set using the following dataset statistics from the initial labeled pool: mean minimum distance between two objects (to set $\varepsilon$) and average area of the bounding boxes (to set $\alpha$). More about selecting these hyperparameters is explained in "Discussion" section. Using the above filtering conditions, the region proposal filtering step effectively retains those region proposals which are likely to enclose some object in the image.

After this step, we have a set of object center clicks and a set of region proposals assigned to each weak label as shown in Fig. 4b. We use their prediction scores to estimate uncertainty.

### *Estimating uncertainty*

We consider sets of filtered proposals for each image from the weak labeled pool. We then estimate uncertainty of a model based on the following hypothesis. If the model is confident and certain about an object, the model's predictions should be invariant to slight changes in the location of bounding boxes. For the sake of illustration, consider one of the five objects present in Fig. 4b. For the set of associated proposals of a given object, a confident model's prediction scores on those proposals would ideally not exhibit high variance. If there exists a high variation in the probabilities of the proposals, the model is deemed to be highly uncertain about that object. We define the following three metrics to estimate our model's uncertainty:

1. *Max-variance (mv)* We calculate variance amongst the prediction scores of the filtered region proposals for each class. For each image $X_i$, we obtain filtered region proposal predictions for a each weak label $w$ in a vector $P_{iw}$. The variance based uncertainty $u_i^{var}$ is calculated as:

$$u_i^{var} = \max_w \frac{1}{|P_{iw}|} \sum_{p \in P_{iw}} (p - mean(P_{iw}))^2 \quad (1)$$

2. *Max-entropy (me)* For each weak label $w$, the sum of entropy at each associated proposal. The entropy based uncertainty $u^{ent}$ for an image is the maximum of such entropy values obtained:

$$u_i^{ent} = \max_w \frac{1}{|P_{iw}|} \sum_{p \in P_{iw}} -p \log_2 p - (1-p) \log_2(1-p) \quad (2)$$

3. *Max-Ent-Var (mev)* We take a linear combination of the above two metrics and use it as an uncertainty metric $u^{ve}$. Max-Variance metric is used as it is but a variant of Max-Entropy is used. For each weak label $w$, we calculate the average of entropy values of its associated proposals. Here $\lambda_1, \lambda_2 \in R$ are hyperparameters and we define the metric as follows:

$$u_i^{ev} = \lambda_1 u_i^{ent} + \lambda_2 u_i^{var} \quad (3)$$

The motivation behind this metric and how the hyperparameters $\lambda_1$ and $\lambda_2$ are chosen is explained in "Discussion" section.

**Algorithm 2:** Proposed Active Sampling Method

**Input:** Weak labeled pool $D^W$, model $\theta$, sample size $b_S$
search radius $\varepsilon$, max region-proposal area $\alpha$
**Output:** An actively sampled images batch of size $b_S$

$U_{image} \leftarrow \{\ \}$
**for** *image i in* $D^W$ **do**
$\quad U \leftarrow \{\ \}$
$\quad$ **for** *each weak label w in* `WeakLabels(i)` **do**
$\quad\quad P_{filtered} \leftarrow \{\ \}$
$\quad\quad$ **for** *each proposal $\rho$ in* `RegionProposals`$(\theta, i)$ **do**
$\quad\quad\quad$ **if** *w present inside $\rho$* **and**
$\quad\quad\quad \|\text{BoxCenter}(\rho) - w\| \leq \varepsilon$ **and** $\text{BoxArea}(\rho) \leq \alpha$ **then**
$\quad\quad\quad\quad$ // Keep the proposal $P_{filtered} \leftarrow P_{filtered} \cup \rho$

$\quad\quad$ // U holds uncertainty of weak label w of image i at [i,w] index
$\quad\quad U[i,w] \leftarrow \text{Uncertainty}(P_{filtered})$
$\quad$ // $U_{image}$ holds the final uncertainty assigned to image i at index i
$\quad U_{image}[i] \leftarrow \max(U[i,:])$
`SortDecreasing`$(U_{image})$    // Sort $U_{image}$ in a decreasing order of uncertainty
**return** images corresponding to $U_{image}[:b_S]$    // Return a batch of uncertain images

As detailed in Algorithm 2, in each episode of the active learning cycle, we query $b_W$ images from the unlabeled pool $D^U$ for Type-1 annotation and move the images to the weakly labeled pool $D^W$. We run the model on $D^W$, which is followed by the RPF step. The region proposals retained after the RPF step are then used to estimate uncertainty of the model on weakly labeled pool ($D^W$) using our methods. We then query $b_S$ images from the weakly labeled pool $D^W$ for Type-2 annotation i.e., the ones the model deems uncertain. Once these images are queried for strong labels, they are moved to the labeled pool $D^L$.

## Experimental setup
### Wheat dataset
This dataset contains high definition images of wheat plants. We refer the reader to Madec et al. [40] for details on data acquisition steps, data preparation procedure and field experiments conducted. To avoid potential storage and computation resource overheads, we preprocessed the original images of size $4000 \times 6000$ to create a dataset suitable for training a deep object detection network. We first down sampled the images by a factor of 2 (to $2000 \times 3000$) using a bi-linear aggregation function. Then, we sliced the down sampled images into image tiles of size $500 \times 500$ with no overlap.[2] Post resizing and slicing if only partial objects are present or if a slice doesn't contain an object at all, we ignore the image entirely to avoid adding potential noise to the model. This preprocessing method was inspired by results reported on the Wheat dataset in Madec et al. [40].

Of the obtained 5506 preprocessed images, we used 3304 (60%) images for active learning and the remaining 2202 (40%) for testing our methods. From the 60% chunk, we start the active learning cycle with just 50 images in the labeled pool. At the beginning of each episode, 50 ($b_W$) images are queried for Type-1 annotation and are moved to weakly labeled pool, of which 25 ($b_S$) most valuable images are queried for Type-2 annotation and are moved to the labeled pool.

### Sorghum dataset
This dataset contains high quality aerial images of Sorghum *(Sorghum bicolor L. Moench)*, a C4 tropical grass that plays an essential role in providing nutrition to humans and livestock, particularly in marginal rainfall environments. We refer the reader to Guo et al. [6] for details on data acquisition steps, data preparation procedure and field experiments conducted. We sliced each original image of size $300 \times 1200$ into four $300 \times 300$ pixel images, with no overlap. After slicing if partial objects are present in a slice, we ignore their respective annotations to avoid adding potential noise.

Of the obtained 4641 preprocessed images, we used 2784 (60%) images for active learning and the remaining 1857 (40%) for testing our methods. From the 60% chunk, we start the active learning cycle with just 50 images in the labeled pool. At the beginning of each episode, 30 ($b_W$) images are queried for Type-1 annotation and are moved to weakly labeled pool, of which 15 ($b_S$) most valuable images are queried for Type-2 annotation and are moved to the labeled pool. The parameter choices for both the datasets are shown in Table 1. Examples of preprocessed images can be seen in Fig. 7.

---

[2] Images and their bounding box annotations were sliced and resized using the library available at: https://image-bbox-slicer.readthedocs.io/.

**Table 1 Hyperparameter choices in methodology implementation**

| Parameters | $D^L$ | $D^U$ | $b_W$ | $b_S$ | $\varepsilon$ | $\alpha$ |
|---|---|---|---|---|---|---|
| Wheat | 50 | 4481 | 50 | 25 | 80 | 20,000 |
| Sorghum | 50 | 3505 | 30 | 15 | 20 | 1400 |

**Implementation details**

All the experiments are conducted with Faster R-CNN [39, 41] as the object detector. The intermediate region proposal layer in Faster R-CNN makes it a natural choice for us and the experiments can be easily extended to any segmentation task which uses Mask-RCNN [42]. The residual network ResNet101 [43] was used as the base network for both the datasets. We trained all our models to minimize the Cross-Entropy loss function with Stochastic Gradient Descent as the optimizer with a learning rate of 0.004 and a mini-batch size of 4 images. The learning rate was decayed every 5 steps by 0.1. After running some initial set of experiments and closely monitoring the loss value trends on both the datasets, we decided to train the models in each active learning cycle for 10 epochs.

We first train a baseline model with a randomly chosen labeled subset of the available data. This model is used as the starting point for active learning. A model trained in a particular cycle is used in the cycle that follows it. As shown in Table 1, we start with a labeled pool $D^L$ and an unlabeled pool $D^U$ and a weakly labeled pool $D^W$. In every cycle, a batch of $b_W$ images from $D^U$ are queried for point supervision and added to $D^W$, then a batch of $b_S$ images from $D^W$ are queried for strong supervision which are added to $D_L$. The images which are queried for point supervision but not queried for strong supervision in every cycle are stored in $D^W$.

**Comparison with baselines**

We compare our proposed methods with the following baselines:

- *Random (rand)* Samples are selected randomly from the unlabeled pool.
- *Least Confident (lc)* Confidence for an image is calculated as the highest bounding box probability in that image. Images with least confidence are selected. This criterion is taken from the min-max method specified in Roy et al. [36].
- *Margin (mar)* For a predicted bounding box, margin is calculated as the difference between probabilities of the top two model predictions. Intuitively, low margin means that the model is uncertain about the data point. For each image, margin is chosen to be the summation of margins across all the predicted bounding boxes in the image. This is taken from Brust et al. [35].
- *Entropy (ent)* Samples with high entropy in the probability distribution of the predictions are selected. This is taken from Roy et al. [36].

Since our proposed methods are two-stage in nature (in the first stage we query images for Type-1 annotation and then for Type-2 in the second stage) we denote them on the result in **{Query_For_Weak}_{Query_For_Strong}** format. So **lc_mv** denotes that images were queried for Type-1 annotation using Least Confidence query method and then in stage two, images are queried for Type-2 annotation based on Max-Variance uncertainty metric described in "Methodology" section.

**Evaluation criteria**

Performance of active learning methods is usually evaluated by plotting a curve between model performance and number of training samples. For each query method, we report every model's *Mean Average Precision* or mAP on a held-out test set against the number of images it was trained on. mAP is the most commonly used evaluation criteria in the object detection space. A predicted bounding box is considered correct (true positive, TP) if it overlaps more than the IOU (intersection-over-union) threshold with a labelled bounding box. Otherwise the predicted bounding box is considered as false positive (FP). When the labelled bounding box have an IOU with a predicted bounding box lower than the threshold value, it is considered as false negative (FN). The standard IOU threshold value of 0.5 was used. The precision and recall are then computed using:

$$Precision = \frac{TP}{TP+FP} \quad \& \quad Recall = \frac{TP}{TP+FN} \quad (4)$$

The score associated to each bounding box allows evaluating the trade-off between false positive and false negative. The average precision (AP@0.5IOU) [44] was used to quantify the detection performances. The standard average precision metrics, is the area under the precision-recall curve obtained for different bounding box scores. The average precision balances the precision and recall performances terms that may be strongly correlated. It varies between 0 (TP = 0) to 1 (FN = 0). We also measure the efficiency of our method by examining the annotation cost estimates. The annotation costs (measured in units of time) are calculated using Eqs. 5 and 6 discussed in the following subsection. We sanction "Results" section of the paper to discuss and validate effectiveness of our proposed active learning methods, which is why we report the best model's crop density performance in "Discussion" section.

### Estimating annotation costs

As the annotation times of the datasets were unavailable, we used statistics of the popular ImageNet dataset for consistency. Su et al. [45] and Papadopoulos et al. [24] report the following median times per image on ImageNet: 25.5 s for drawing one box, 9.0 s for verifying its quality and 7.8 s for checking whether there are other objects in the image yet to be annotated and 3.0 s to click on an object's center. Taking these into account, we calculate that Type-1 annotation (object clicking + checking whether there are other objects) requires 10.8 s. And Type-2 annotation requires 34.5s, 7.8s less than how much traditional bounding box annotations take since there is no need to check whether there are other object in the image. So for baseline methods, given a batch of queried images of size $Q$, with a total of $b_Q$ objects in it, we calculate annotation time (in seconds) using the following formula:

$$Time = 7.8 \times Q + 34.5 \times b_Q \quad (5)$$

In case of our proposed methods, given a images batch of size $Q_W$, with a total of $b_{QW}$ objects queried for Type-1 annotations and a batch of size $Q_S$, with a total of $b_{QS}$ objects queried for Type-2 annotation, we calculate annotation time using the following formula:

$$Time = 7.8 \times Q_W + 34.5 \times b_{QS} + 3 \times b_{QW} \quad (6)$$

### Results

#### Results on Wheat

Figure 5a shows how test mAP increased with the number of training examples, Fig. 5b compares annotation costs (time in hours) incurred by our methods with respect to the annotation cost incurred by the best standard PBAL (Pool Based Active Learning) baseline method (black dashed line). From the plot, it is clear that after just 2 episodes all our methods start to maintain a higher mAP compared to that of the baselines. The best baseline is the entropy based standard PBAL (ent) with 0.7631 mAP, which required the oracle to label 900 images, costing 29.14 hours of annotation. As shown in Fig. 5b, 3 variants of the **Max Ent-Var** method (mar_mev, lc_mev and ent_mev) outperform the best baseline method with approximately 60% lesser images (350) costing the oracle approximately 50% lesser annotation time (10.52, 11.13 and 12.72 hours respectively). Object detectors trained using all variants of our proposed methods have performances better than those of the best baseline method at the end of the active learning episodes (900 images).

#### Results on Sorghum

Similar to wheat plots, Fig. 6a shows how test mAP increased with the number of training examples, Fig. 6b compares annotation costs (time in hours) incurred by our methods with respect to the annotation cost incurred by the best standard PBAL baseline method (black dashed line). Unlike in the case of Wheat, object detector's performance on Sorghum improves steeply in the beginning. Even in the case of Sorghum, the best baseline is the entropy based standard PBAL (ent) with 0.8136 mAP, which required the oracle to label 500 images, costing 106.76 h of annotation. As shown in Fig. 6b, 6 of the proposed methods outperform the best baseline method with less than 60 h of annotation costs, which is more than 55% in savings when compared to cost incurred by the best baseline method. Object detectors trained using 10 out of 12 variants of our proposed methods have performances better than that of the best baseline method at the end of the active learning episodes.

### Discussion

#### Analysis of *most valuable* images

To better understand the performance of our proposed methods, we observe the kinds of images queried across different episodes of active learning. Figure 7 shows the most informative images sampled using Max-Ent-Var method in episodes 1, 2 and 3. By observing the queried samples, we can see that in the first episode, the model is uncertain about images with objects which are either blurred, occluded or in bad lighting conditions. For images containing such adversarial features, this behaviour intuitively makes sense as it is often hard for the model to find even simple patterns like edges and corners and thus the prediction variance is naturally expected to be high for them. In episode 2, although images with blurred objects are queried, our model also sampled images with bad lightning conditions. In episode 3 on wheat, all the top 5 sampled images have dry leaves. We believe the reason for this might be the fact that it is difficult to detect wheat ears with dry background. In case of Sorghum, sampled images in episode 3 indicate that the model is struggling to correctly detect and localize objects in bright light where the sorghum head is almost white in colour.

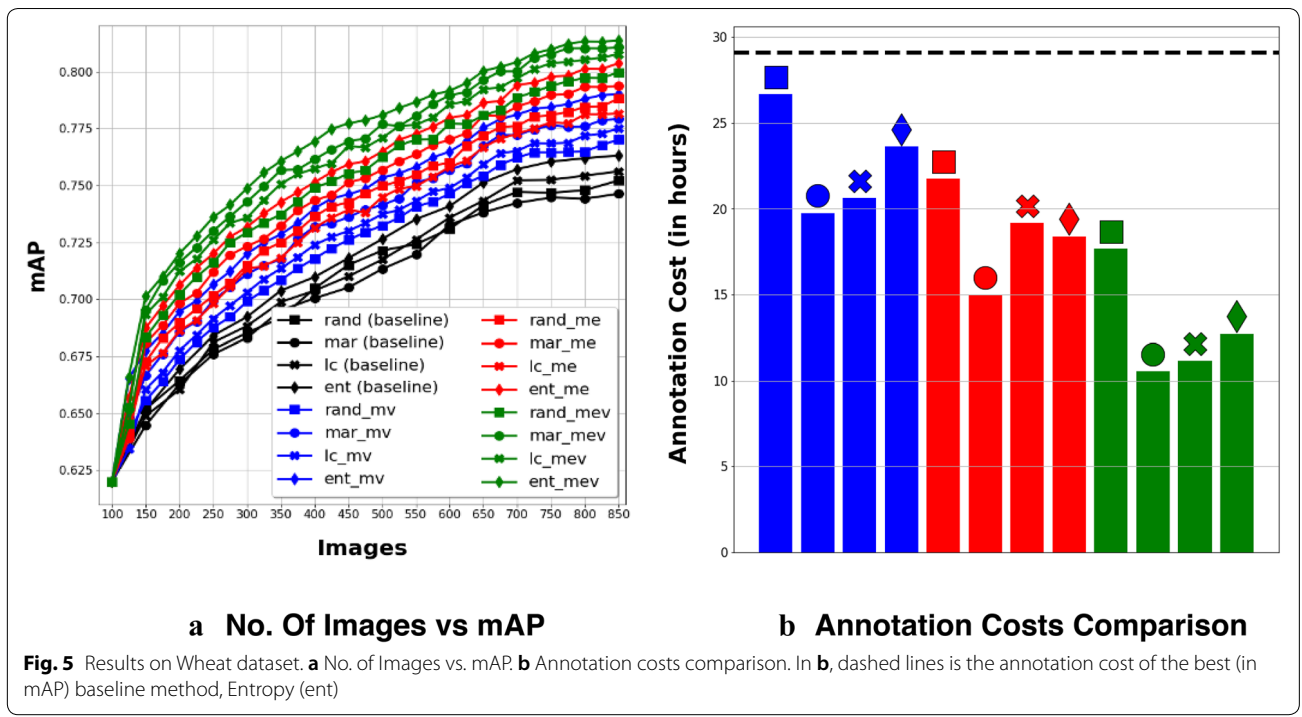

**Fig. 5** Results on Wheat dataset. **a** No. of Images vs. mAP. **b** Annotation costs comparison. In **b**, dashed lines is the annotation cost of the best (in mAP) baseline method, Entropy (ent)

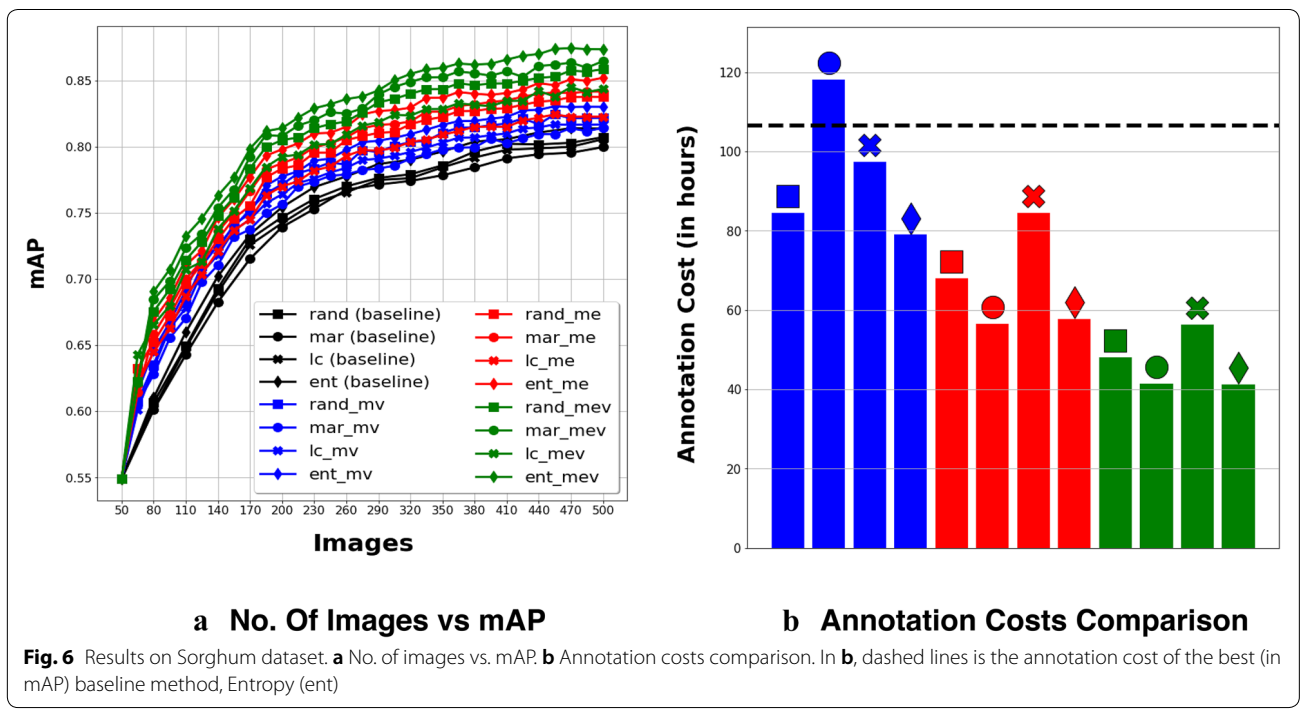

**Fig. 6** Results on Sorghum dataset. **a** No. of images vs. mAP. **b** Annotation costs comparison. In **b**, dashed lines is the annotation cost of the best (in mAP) baseline method, Entropy (ent)

## On similarities between region proposal filtering and non-maximum suppression

Non-maximum suppression (NMS) has been widely used in several key aspects of computer vision. It is an integral part of tasks such as edge, corner and object detection

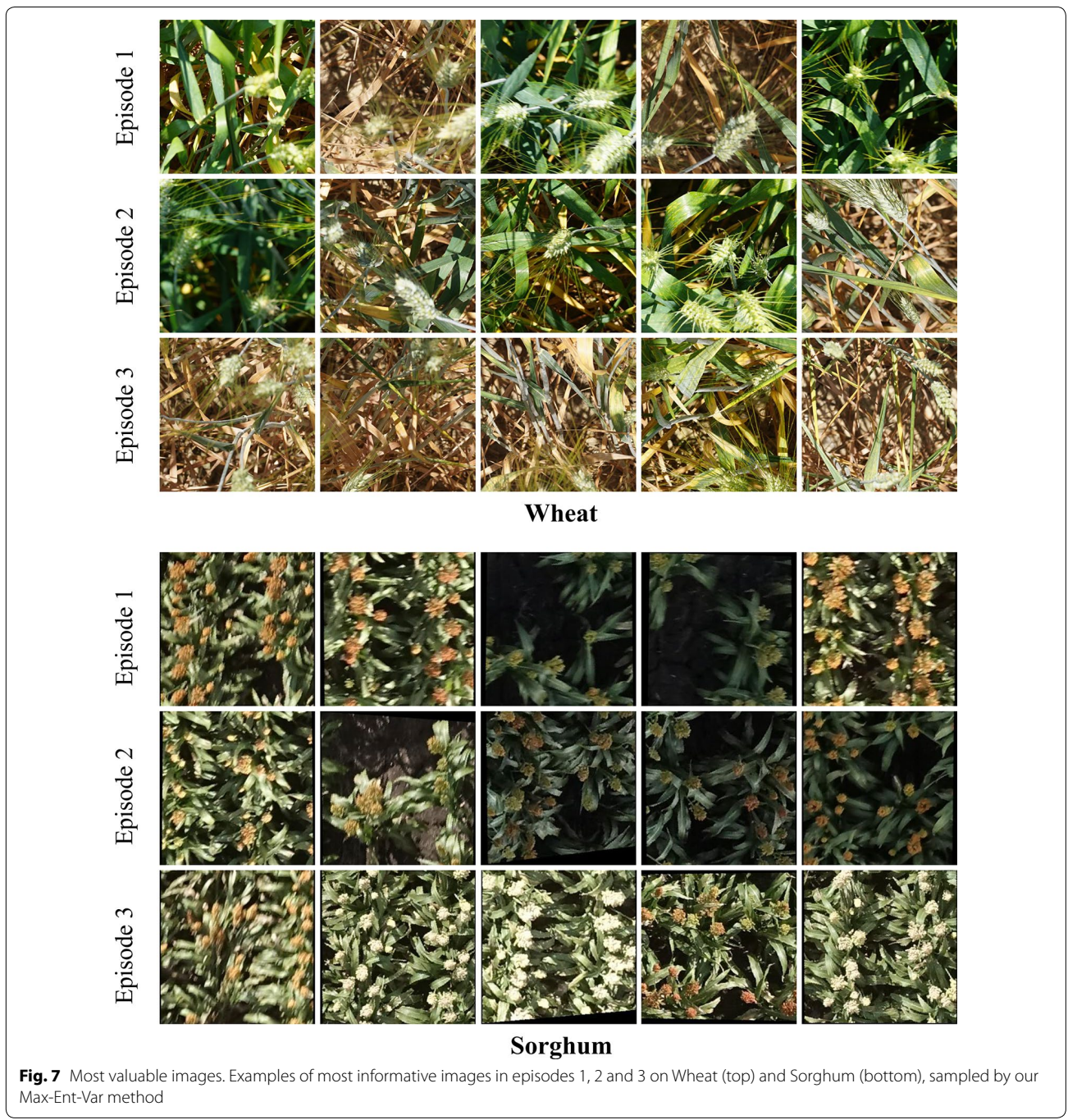

**Fig. 7** Most valuable images. Examples of most informative images in episodes 1, 2 and 3 on Wheat (top) and Sorghum (bottom), sampled by our Max-Ent-Var method

[46–50]. In the context of object detection, the apparent similarity between NMS and Region Proposal Filtering (RPF) is that they both filter out unwanted region proposals. NMS removes the region proposals which highly overlap the region proposal with maximum confidence. Its effect can be controlled by specifying the overlapping criterion, intersection-over-union (IoU) which is a hyperparameter. The goal of NMS therefore is to retain only one region proposal per each prediction group, corresponding to the precise local maximum of the model's output, ideally obtaining only one bounding box per object. The motivation behind RPF is quite different. In active learning, a sampling function samples the most informative data points for labeling. An uncertainty sampling function [31] essentially quantifies the informativeness based on how uncertain the model predictions are.

This motivates our RPF method, where we aim to keep region proposals that overlap each other by a great deal but also vary slightly, in contrast with NMS. We utilize RPF to validate our hypothesis that a detection model's predictions on an object can be considered uncertain if they significantly vary with a slight change in the object's environment i.e., image areas immediately surrounding the object. In other words, we quantify uncertainty by estimating the variance in the model's confidence values for similar looking proposals. Our experimental results and ablation studies support this hypothesis and show that the model queries images with blurry object backgrounds and other occlusions in the early stages of training. Also, the model generalizes faster with fewer training data points when proposed RPF method is used.

### Tuning hyperparameters for region proposal filtering

Region Proposal Filtering (RPF), shown in Fig. 4, is a novel and a very crucial step in our proposed methodology and there are two hyper-parameters $\varepsilon$ and $\alpha$ to be tuned to make it work the best. We select the value for $\varepsilon$ by examining (a) the distribution of minimum distance between any two objects (bounding box centers) and select the value for $\alpha$ by observing (b) the distribution of area of the boxes. To avoid the problem of having two objects in a same region proposal at the same time, post RPF step, we pick the 20th percentile of (a) as $\varepsilon$ for both datasets. The 20th percentile of (a) for the Wheat dataset was 18 so we rounded it to 20, for the Sorghum dataset it was close to 77 so we rounded it to 80. Statistically with these values, after the RPF step, filtered region proposals will not have a second object in the image 80% of the time. We believe this is robust enough since by default, RPF adds an extra filter by dropping all proposals that contain other weak labels, other than the weak-label-of-interest. With similar motivation, we pick 90th percentile of (b) for both datasets which suggests that statistically, just 10% of the time after the RPF step, filtered region proposals will not include the actual bounding box. The 90th percentile of (b) for Wheat was 20,448, rounded to 20,000, for Sorghum it was 1404, rounded to 1400.

### Choosing $\lambda_1$, $\lambda_2$ in Max-Ent-Var (mev)

The performance of our best active sampling method, Max-Ent-Var (mev), is highly dependant on $\lambda_1$, $\lambda_2$ hyperparameters. In our attempt to combine two quantities (entropy and variance) which have different ranges, different motivation, basically different roles to perform. We theoretically and empirically examine the upper bounds of both quantities in an effort to combine them efficiently. The motivation to combine them is explained with the following toy example: Imagine that the following are the probabilities of region proposals left around a particular object after RPF layer—[0.5, 0.5, 0.5, 0.5, 0.5]. Clearly, the model is highly uncertain about the object so our entropy based metric outputs its maximum value (for this object) of 1, which is ideal. But our variance based metric Max-Variance (mv) outputs its minimum value of 0 which indicates that the model is highly certain about the object. To overcome these rarely occurring shortfalls of our methods, we decided to come up with a linear combination of both the metrics.

To combine both entropy and variance metrics, we first make sure they are on similar scales i.e., have similar minima and maxima. In case of entropy, the theory suggests that the range of entropy of a distribution with $n$ number of outcomes is given by:

$$0 \leq Entropy \leq \log_2(n) \quad (7)$$

In our case $n$ is 2 (object or not) so we can say that $u_i^{ent} \in [0, 1]$. In case of variance, we use the Bhatia–Davis inequality [51] to find the upper bound on variance of values from a known distribution. Suppose a distribution has minimum $m$, maximum $M$, and expected value $\mu$. Then the inequality says:

$$\sigma^2 \leq (M - \mu)(\mu - m) \quad (8)$$

In our case, $m$ is 0, $M$ is 1 and we can safely assume the worst case of $\mu$ as 0.5. Plugging in the values into Eq. 8, we get that $u_i^{var} \in [0, 0.25]$. So the most straightforward thing to do here to bring them to same scale would be to multiply $u_i^{var}$ by 4.

Figure 8 includes empirical analysis of these metrics. In Fig. 8a, you can see the distribution of $u_i^{ent}$. Both original and scaled distributions of $u_i^{var}$ can be seen in Fig. 8b, c respectively. These values calculated during the active learning cycle on Wheat dataset. Figure 8c reassures, empirically, that scaling $u_i^{var}$ by 4 indeed makes its feasible to add it with $u_i^{ent}$ as they have similar ranges and similar contribution to the $u^{ev}$ metric. Figure 8e shows a scatter plot between corresponding $u_i^{ent}$ and scaled $u_i^{var}$ values and it is interesting to see a pattern appear between them where after an apparent threshold value, the both values are never simultaneously high. This explains why the $u_i^{ev}$ has almost a perfect normal distribution centered at 0.7, as shown in Fig. 8d.

### Motivation for margin and entropy sampling methods

We refer the readers to section 3.1 of Settles [31] where it has been summarized why margin and entropy based sampling methods are often favoured by active learning practitioners. We merely reiterate the same explanation and illustrate a simple example here to further convince the readers on the benefits of these methods. By definition, the least confidence method uses only the maximum

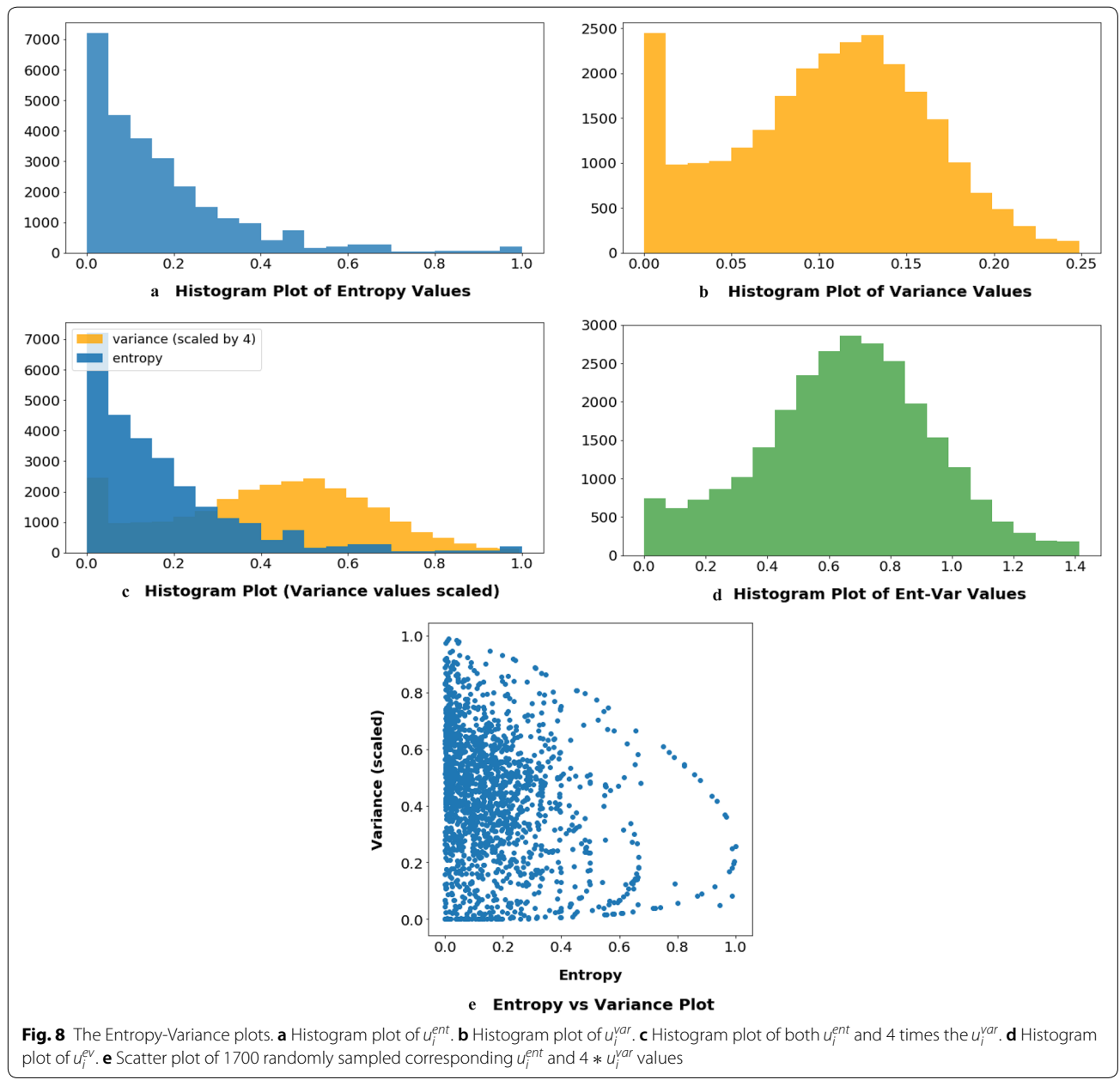

**Fig. 8** The Entropy-Variance plots. **a** Histogram plot of $u_i^{ent}$. **b** Histogram plot of $u_i^{var}$. **c** Histogram plot of both $u_i^{ent}$ and 4 times the $u_i^{var}$. **d** Histogram plot of $u_i^{ev}$. **e** Scatter plot of 1700 randomly sampled corresponding $u_i^{ent}$ and $4*u_i^{var}$ values

probability value to calculate model's confidence/uncertainty value and *throws away* information available in the rest of the distribution. Considering $x$ as input, $y$ as target labels, $\hat{y}$ as predicted labels and $x^*$ as the most informative instance (i.e., the best query), least confidence is calculated as follows:

$$x_{LC}^* = \arg\max_x 1 - P_\theta(\hat{y}|x) \qquad (9)$$

where $\hat{y} = \arg\max_y P_\theta(y|x)$, or the class label with the highest posterior probability under the model $\theta$. To correct the shortcomings in least confident strategy, margin based sampling method upgrades slightly on the least confidence method by using more information - the probabilities of top two labels. To calculate model's uncertainty, margin sampling takes the difference between the probabilities of first and second most probable labels as follows:

$$x_M^* = \arg\min_x P_\theta(\hat{y}_1|x) - P_\theta(\hat{y}_2|x) \qquad (10)$$

where $\hat{y}_1$ and $\hat{y}_2$ are the first and second most probable class labels. The lower the margin, the higher the uncertainty of the model thus knowing the true label would help the model discriminate more effectively between them. Entropy is a more general uncertainty sampling strategy that uses the entire distribution:

$$x_H^* = \arg\max_x -\sum_i P_\theta(y_i|x) \log P_\theta(y_i|x) \quad (11)$$

where $y_i$ ranges over all possible labels. Entropy, an information-theoretic measure that represents the amount of information needed to encode a distribution, is thought of as a measure of uncertainty in machine learning.

Consider a 5-class classification task, a trained model's three output probability distributions could be $\hat{y}^1 = [0.05, 0.5, 0.2, 0.05, 0.2]$, $\hat{y}^2 = [0.02, 0.5, 0.2, 0.03, 0.15]$ and $\hat{y}^3 = [0.1, 0.5, 0.2, 0.1, 0.1]$. These seemingly similar distributions are quite different from each other in the context of uncertainty. Least confidence and margin sampling, as illustrated in Table 2, find it hard to discriminate these particular examples as the calculated uncertainty values are same for all. For these examples, least confident and margin sampling methods are no better than random. On the other hand, entropy takes advantage of the entire distribution to overcome the extant shortcomings to query $\hat{y}^3$.

### Verifying crop density performance

Throughout the paper, we evaluated detection models learned using our methodology on the basis of their Mean Average Precision (mAP) but here we also evaluate our model on one of the tasks that measures grain yield-crop density estimation. We see that the models trained on our best method Max-Ent-Var (mve) show exceptional performance in crop density estimation task, evaluated on Pearson's Correlation Coefficient and Root Mean Squared Error. Pearson's correlation coefficient, commonly represented by $r$, is a measure of how similar two data distribution are. Given paired data $\{\{x_1, y_1\}, \ldots, \{x_n, y_n\}\}$ consisting of $n$ pairs, $r_{xy}$ is defined as:

$$r_{xy} = \frac{\sum_{i=1}^n (x_i - \bar{x})(y_i - \bar{y})}{\sqrt{\sum_{i=1}^n (x_i - \bar{x})^2}\sqrt{\sum_{i=1}^n (y_i - \bar{y})^2}} \quad (12)$$

Where $n$ is sample size, $x_i, y_i$ are the individual sample points indexed with i, $\bar{x} = \frac{1}{n}\sum_{i=1}^n x_i$ is the mean of all $x$ values and analogously for $\bar{y}$. The root-mean-squared deviation (RMSD) or root-mean-squared error (RMSE) is a common measure of the differences between values predicted by a model and the values observed. Given the same pair of values mentioned before Eq. 12, the formula to calculate RMSE is:

$$\text{RMSE} = \sqrt{\frac{1}{n}\sum_{i=1}^n (x_i - y_i)^2} \quad (13)$$

Figure 9 shows the scatter plots between model predicted crop count and actual crop count of models trained on both Wheat and Sorghum datasets in Fig. 9a, b respectively. The plots include correlation coefficients and RMSE values at the bottom. In case of Wheat, the Pearson's correlation coefficient ($r$) is 0.8147 while RMSE is 1.2080. In case of Sorghum, $r$ is 0.9097 (high) while RMSE is 2.7069. We report these results on a large test sets—2202 images in case of Wheat and 1857 images in case of Sorghum.

### Extensibility to multi class datasets

In our experiments, we report results on only single class detection datasets. However, our methods can be readily applied on multi-class datasets because the concepts of entropy and variance can be easily generalized to work with multiple classes. In the future, we hope to extend our work to multi-class detection datasets. Since the datasets in our current experiments have a single class, we simply use the *objectness scores* of the region proposal network (RPN) as the bounding box predictions in our implementation. Methodology wise, the same technique can be seamlessly extended to multi-class detection datasets by instead looking at the final output vector. This is because the objectness scores alone may not be the best estimator for the uncertainty of the model.

### Other forms of weak supervision

After running preliminary experiments on VOC [37], a dataset with 20 classes, we found that our methodology works decently when provided with localization based weak signals (object center clicks) but doesn't work well with a much more affordable image level weak signal i.e., in the Type-1 annotation step, the annotator only provides classes of objects present in images as weak labels. In our future work, we will compare the effect of various forms of weak supervision on the active learning process.

**Table 2 Uncertainty sampling example**

| Sampling method | $\hat{y}^1$ | $\hat{y}^2$ | $\hat{y}^3$ | Queried sample |
|---|---|---|---|---|
| Least confidence | 1−0.5 = 0.5 | 1−0.5 = 0.5 | 1−0.5 = 0.5 | $\hat{y}_1$ or $\hat{y}_2$ or $\hat{y}_3$ |
| Margin | 0.5−0.2 = 0.3 | 0.5−0.2 = 0.3 | 0.5−0.2 = 0.3 | $\hat{y}_1$ or $\hat{y}_2$ or $\hat{y}_3$ |
| Entropy | 1.86 | 1.66 | 1.96 | $\hat{y}_3$ |

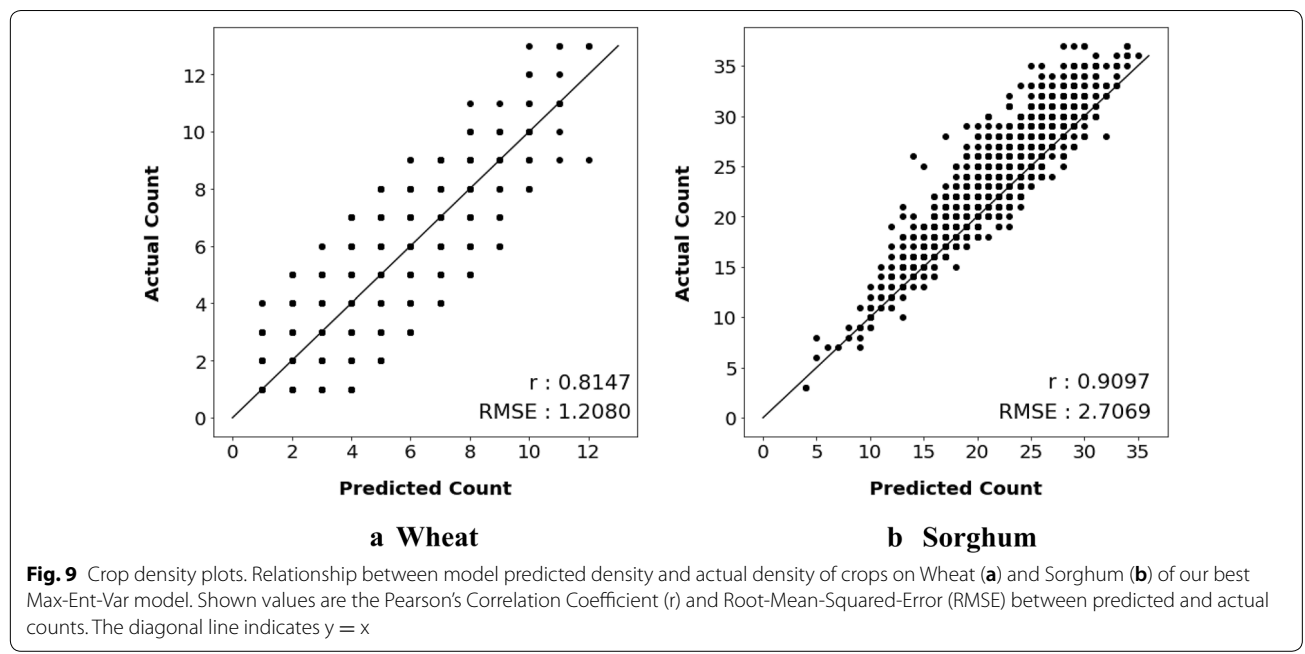

**Fig. 9** Crop density plots. Relationship between model predicted density and actual density of crops on Wheat (**a**) and Sorghum (**b**) of our best Max-Ent-Var model. Shown values are the Pearson's Correlation Coefficient (r) and Root-Mean-Squared-Error (RMSE) between predicted and actual counts. The diagonal line indicates y = x

## Conclusion

The methodology described in this work demonstrates a novel, effective active learning method for object detection in crop images based on point supervision. By performing extensive experiments on Sorghum and Wheat datasets, we have empirically shown that point supervision significantly improves the query performance by picking highly informative images. Our qualitative results also reinforce the phenomenon that querying with our proposed method results in picking blurry and low light images in which the objects are intuitively harder to localize accurately. This behavior is highly desirable in the case of object detection in crop images because crop datasets often have images with blur, occlusion and bad lighting conditions. The proposed active learning framework can be extended to incorporate bounding box size information for uncertainty estimation. Also, a natural extension to the proposed method is to make RPF parameters learnable. We leave this for future work.

#### Acknowledgements
We thank the anonymous reviewers for their valuable comments and suggestions.

#### Authors' contributions
Both ALC and SVD proposed the methods, analyzed the data, designed the experiments, and interpreted results while VNB, WG, and SN supervised the entire study. ALC implemented the proposed methods in code, conducted ablation studies required and prepared necessary figures. Both ALC and SVD wrote the paper with input from all authors. All authors read and approved the final manuscript.

#### Funding
This study was partially funded by Indo-Japan DST-JST SICORP program "Data Science-based Farming Support System for Sustainable Crop Production under Climatic Change" and CREST Program "Knowledge Discovery by Constructing AgriBigData" (JPMJCR1512) from Japan Science and Technology Agency.

#### Data availability statement
The datasets and materials will be provided on publication

#### Ethics approval and consent to participate
Not applicable.

#### Consent for publication
Not applicable.

#### Competing interests
The authors declare that they have no competing interests.

Received: 6 October 2019   Accepted: 22 February 2020
Published online: 07 March 2020

### References
1. Grinblat GL, Uzal LC, Larese MG, Granitto PM. Deep learning for plant identification using vein morphological patterns. Comput Electron Agric. 2016;127:418–24. https://doi.org/10.1016/j.compag.2016.07.003.
2. Ghosal S, Blystone D, Singh AK, Ganapathysubramanian B, Singh A, Sarkar S. An explainable deep machine vision framework for plant stress phenotyping. Proc Natl Acad Sci. 2018;115(18):4613–8. https://doi.org/10.1073/pnas.1716999115.
3. Ghosal S, Zheng B, Chapman SC, Potgieter AB, Jordan DR, Wang X, Singh AK, Singh A, Hirafuji M, Ninomiya S, Ganapathysubramanian B, Sarkar S, Guo W. A weakly supervised deep learning framework for sorghum head detection and counting. Plant Phenomics. 2019;2019:1–14. https://doi.org/10.34133/2019/1525874.
4. Desai SV, Balasubramanian VN, Fukatsu T, Ninomiya S, Guo W. Automatic estimation of heading date of paddy rice using deep learning. Plant Methods. 2019;15(1):76. https://doi.org/10.1186/s13007-019-0457-1.
5. Hasan MM, Chopin JP, Laga H, Miklavcic SJ. Detection and analysis of wheat spikes using convolutional neural networks. Plant Methods. 2018;14(1):100. https://doi.org/10.1186/s13007-018-0366-8.

6. Guo W, Zheng B, Potgieter AB, Diot J, Watanabe K, Noshita K, Jordan DR, Wang X, Watson J, Ninomiya S, Chapman SC. Aerial imagery analysis—quantifying appearance and number of sorghum heads for applications in breeding and agronomy. Front Plant Sci. 2018;9:1544. https://doi.org/10.3389/fpls.2018.01544.
7. Sadeghi-Tehran P, Virlet N, Ampe EM, Reyns P, Hawkesford MJ. Deepcount: in-field automatic quantification of wheat spikes using simple linear iterative clustering and deep convolutional neural networks. Front Plant Sci. 2019;10:1176. https://doi.org/10.3389/fpls.2019.01176.
8. Ubbens J, Cieslak M, Prusinkiewicz P, Stavness I. The use of plant models in deep learning: an application to leaf counting in rosette plants. Plant Methods. 2018;14:6.
9. Sa I, Ge Z, Dayoub F, Upcroft B, Perez T, McCool C. Deepfruits: a fruit detection system using deep neural networks. Sensors. 2016;16:1222.
10. Xiong X, Duan L, Liu L, Tu H, Yang P, Wu D, Chen G, Xiong L, Yang W, Liu Q. Panicle-seg: a robust image segmentation method for rice panicles in the field based on deep learning and superpixel optimization. Plant Methods. 2017;13(1):104. https://doi.org/10.1186/s13007-017-0254-7.
11. Oh M-h, Olsen P, Ramamurthy KN. Counting and segmenting sorghum heads. 2019. arXiv:1905.13291
12. Milioto A, Lottes P, Stachniss C. Real-time semantic segmentation of crop and weed for precision agriculture robots leveraging background knowledge in CNNs. In: 2018 IEEE international conference on robotics and automation (ICRA). 2018. p. 2229–35.
13. Kamilaris A, Prenafeta-Boldú FX. Deep learning in agriculture: a survey. Comput Electron Agric. 2018;147:70–90. https://doi.org/10.1016/j.compag.2018.02.016.
14. Singh AK, Ganapathysubramanian B, Sarkar S, Singh A. Deep learning for plant stress phenotyping: trends and future perspectives. Trends Plant Sci. 2018;23(10):883–98. https://doi.org/10.1016/j.tplants.2018.07.004.
15. Albarqouni S, Baur C, Achilles F, Belagiannis V, Demirci S, Navab N. Aggnet: deep learning from crowds for mitosis detection in breast cancer histology images. IEEE Trans Med Imaging. 2016;35(5):1313–21. https://doi.org/10.1109/TMI.2016.2528120.
16. Russakovsky O, Li L, Fei-Fei L. Best of both worlds: human–machine collaboration for object annotation. In: 2015 IEEE conference on computer vision and pattern recognition (CVPR). 2015. p. 2121–31. https://doi.org/10.1109/CVPR.2015.7298824
17. Papadopoulos DP, Uijlings JRR, Keller F, Ferrari V. We don't need no bounding-boxes: training object class detectors using only human verification. CoRR. 2016. arXiv:abs/1602.08405.
18. Vijayanarasimhan S, Grauman K. Large-scale live active learning: training object detectors with crawled data and crowds. Int J Comput Vis. 2014;108(1):97–114. https://doi.org/10.1007/s11263-014-0721-9.
19. Yao A, Gall J, Leistner C, Van Gool L. Interactive object detection. In: 2012 IEEE conference on computer vision and pattern recognition. 2012. p. 3242–9. https://doi.org/10.1109/CVPR.2012.6248060
20. Barsoum E, Zhang C, Ferrer CC, Zhang Z. Training deep networks for facial expression recognition with crowd-sourced label distribution. In: Proceedings of the 18th ACM international conference on multimodal interaction. ICMI '16. New York: Association for Computing Machinery; 2016. p. 279–83. https://doi.org/10.1145/2993148.2993165.
21. Kawano Y, Yanai K. Automatic expansion of a food image dataset leveraging existing categories with domain adaptation. In: Agapito L, Bronstein MM, Rother C, editors. Computer vision–ECCV 2014 workshops. Cham: Springer; 2015. p. 3–17.
22. Welinder P, Perona P. Online crowdsourcing: Rating annotators and obtaining cost-effective labels. In: 2010 IEEE computer society conference on computer vision and pattern recognition—workshops. 2010. p. 25–32. https://doi.org/10.1109/CVPRW.2010.5543189
23. Sorokin A, Forsyth D. Utility data annotation with amazon mechanical turk. In: 2008 IEEE computer society conference on computer vision and pattern recognition workshops. 2008. p. 1–8. https://doi.org/10.1109/CVPRW.2008.4562953
24. Papadopoulos DP, Uijlings JRR, Keller F, Ferrari V. Training object class detectors with click supervision. CoRR. 2017. arXiv:abs/1704.06189.
25. Papadopoulos DP, Uijlings JRR, Keller F, Ferrari V. Extreme clicking for efficient object annotation. In: 2017 IEEE international conference on computer vision (ICCV), 2017. p. 4940–9. https://doi.org/10.1109/ICCV.2017.528
26. Russakovsky O, Bearman AL, Ferrari V, Li F. What's the point: semantic segmentation with point supervision. CoRR. 2015. arXiv:abs/1506.02106.
27. Mettes P, Gemert J, Snoek C. Spot on: Action localization from pointly-supervised proposals. 2016.
28. Teng E, Falcão JD, Iannucci B. Clickbait: click-based accelerated incremental training of convolutional neural networks. CoRR. 2017. arXiv:abs/1709.05021.
29. Bilen H, Vedaldi A. Weakly supervised deep detection networks. In: 2016 IEEE conference on computer vision and pattern recognition (CVPR). 2015. p. 2846–54.
30. Papadopoulos DP, Uijlings JRR, Keller F, Ferrari V. We don't need no bounding-boxes: training object class detectors using only human verification. In: 2016 IEEE conference on computer vision and pattern recognition (CVPR). 2016. p. 854–63.
31. Settles B. Active learning literature survey. Technical report, University of Wisconsin-Madison. 2010.
32. Gal Y, Islam R, Ghahramani Z. Deep bayesian active learning with image data. In: ICML. 2017.
33. Sener O, Savarese S. Active learning for convolutional neural networks: a core-set approach. In: ICLR 2018. 2018.
34. Wang K, Zhang D, Li Y, Zhang R, Lin L. Cost-effective active learning for deep image classification. IEEE Trans Circuit Syst Video Technol. 2017;27(12):2591–600. https://doi.org/10.1109/TCSVT.2016.2589879.
35. Brust C, Käding C, Denzler J. Active learning for deep object detection. CoRR. 2018. arXiv:abs/1809.09875.
36. Roy S, Unmesh A, Namboodiri VP. Deep active learning for object detection. In: BMVC. 2018.
37. Everingham M, Eslami SMA, Van Gool L, Williams CKI, Winn J, Zisserman A. The pascal visual object classes challenge: a retrospective. Int J Comput Vis. 2015;111(1):98–136.
38. Lin T, Maire M, Belongie SJ, Bourdev LD, Girshick RB, Hays J, Perona P, Ramanan D, Dollár P, Zitnick CL. Microsoft COCO: common objects in context. CoRR. 2014. arXiv:abs/1405.0312.
39. Ren S, He K, Girshick RB, Sun J. Faster R-CNN: towards real-time object detection with region proposal networks. CoRR. 2015. arXiv:abs/1506.01497.
40. Madec S, Jin X, Lu H, de Solan B, Liu S, Duyme F, Heritier E, Frederic B. Ear density estimation from high resolution RGB imagery using deep learning technique. Agric For Meteorol. 2019;264:225–34. https://doi.org/10.1016/j.agrformet.2018.10.013.
41. Yang J, Lu J, Batra D, Parikh D. A faster pytorch implementation of faster r-cnn. 2017. https://github.com/jwyang/faster-rcnn.pytorch.
42. He K, Gkioxari G, Dollár P, Girshick RB. Mask r-cnn. In: 2017 IEEE international conference on computer vision (ICCV). 2017. p. 2980–8.
43. He K, Zhang X, Ren S, Sun J. Deep residual learning for image recognition. In: 2016 IEEE conference on computer vision and pattern recognition (CVPR). 2016. p. 770–8.
44. Salton G, McGill MJ. Introduction to modern information retrieval. New York: McGraw-Hill Inc; 1986.
45. Su H, Deng J, Fei-Fei L. Crowdsourcing annotations for visual object detection. AAAI workshops. 2012.
46. Canny J. A computational approach to edge detection. IEEE Trans Pattern Anal Mach Intell. 1986;8:679–98. https://doi.org/10.1109/TPAMI.1986.4767851.
47. Girshick R, Donahue J, Darrell T, Malik J. Rich feature hierarchies for accurate object detection and semantic segmentation. 2013. arXiv:1311.2524.
48. Forsyth D. Object detection with discriminatively trained part-based models. Computer. 2014;47:6–7. https://doi.org/10.1109/MC.2014.42.
49. Dalal N, Triggs B. Histograms of oriented gradients for human detection, vol. 1. 2005. p. 886–93. https://doi.org/10.1109/CVPR.2005.177.
50. Viola P, Jones M. Robust real-time object detection, vol. 57. 2001.
51. Bhatia R, Davis C. A better bound on the variance. Am Math Monthly. 2000;107(4):353–7.

## Publisher's Note